\documentclass[runningheads]{llncs}
\usepackage{graphicx}
\usepackage{amsmath,amssymb} 
\usepackage{color}
\usepackage[width=122mm,left=12mm,paperwidth=146mm,height=193mm,top=12mm,paperheight=217mm]{geometry}
\usepackage{algorithm,algorithmicx,algpseudocode,listings}        
\usepackage{booktabs}
\usepackage{array}
\usepackage{subcaption}
\usepackage{stackengine}
\usepackage{hyperref}

\newcommand\nnfootnote[1]{%
  \begin{NoHyper}
  \renewcommand\thefootnote{}\footnote{#1}%
  \addtocounter{footnote}{-1}%
  \end{NoHyper}
}
\newcolumntype{C}[1]{>{\centering}m{#1}}
\DeclareMathOperator*{\argmin}{arg\,min}
\begin{document}
\pagestyle{headings}
\mainmatter
\def\ECCV18SubNumber{2739}  

\title{Realtime Time Synchronized Event-based Stereo}

\titlerunning{Realtime Time Synchronized Event-based Stereo}
\author{Alex Zihao Zhu\orcidID{0000-0002-2195-014X} \and Yibo Chen\orcidID{0000-0001-9542-7741} \and Kostas Daniilidis\orcidID{0000-0003-0498-0758}}
\authorrunning{A. Z. Zhu, Y. Chen, K. Daniilidis}

\institute{University of Pennsylvania, Philadelphia PA 19104, USA}
\maketitle
\begin{abstract}
In this work, we propose a novel event based stereo method which addresses the problem of motion blur for a moving event camera. Our method uses the velocity of the camera and a range of disparities to synchronize the positions of the events, as if they were captured at a single point in time. We represent these events using a pair of novel time synchronized event disparity volumes, which we show remove motion blur for pixels at the correct disparity in the volume, while further blurring pixels at the wrong disparity. We then apply a novel matching cost over these time synchronized event disparity volumes, which both rewards similarity between the volumes while penalizing blurriness. We show that our method outperforms more expensive, smoothing based event stereo methods, by evaluating on the Multi Vehicle Stereo Event Camera dataset.
\end{abstract}
\nnfootnote{Supplementary video: \url{https://youtu.be/4oa7e4hsrYo}.}
\section{Introduction}
Event cameras are neuromorphically inspired asynchronous visual sensors that register changes in the log intensity of the image. When such a change is detected, the camera immediately returns an event, $(x,y,t,p)$, to the host, consisting of the pixel position of the change, $x,y$, timestamp, $t$, accurate to microseconds, and polarity, $p\in \{-1, 1\}$, indicating whether the intensity decreased or increased. Over time, the output of the camera can be represented as a constant stream of events. The asynchronous nature of the cameras, combined with   with extremely high temporal resolution, allow for high speed, low latency measurements, in situations where traditional cameras may fail. In addition, the cameras exhibit very high dynamic range (120dB vs 60dB for traditional cameras), allowing them to operate in a number of challenging lighting environments. Finally, the cameras also have much lower bandwidth and power consumption. One interesting use case for these cameras is stereo depth estimation, where they can provide high speed depth information for tasks such as obstacle avoidance and high speed 3D tracking. 

However, a major problem facing general event-based methods is that of time synchronization. That is, events generated at different times may correspond to the same point in the image space, but will appear at different pixel positions due to the motion of the point. This problem manifests itself in two ways. Between cameras, this problem is analogous to having unsynchronized cameras for frame based stereo methods, where the epipolar constraint breaks down due to the motion between the images, and occurs when events are not generated at the same time between the two cameras. Within a single camera, this causes effects similar to the motion blur seen in frame based images. For the stereo matching problem, which is often solved using appearance based similarity metrics, this blurring is highly detrimental, as it often alters the appearance of each image differently. A number of event based stereo methods have approached these problem with asynchronous methods (e.g. \cite{piatkowska2014cooperative,rogister2012asynchronous,xie2017event}), which process each event independently. However, these methods must either forego the information provided by the spatial neighborhood around each event, or use fine tuned temporal windows to once again ensure time synchronization, as there are no guarantees that neighboring events were generated at a similar time.


In this work, we show that this problem can be resolved for stereo disparity matching if the velocity of the camera is known. In particular, we propose a novel event disparity volume for events from a stereo event camera pair, that uses the motion of the camera to temporally synchronize the events with temporal interpolation at each disparity. Our method takes inspiration from past works from Zhu et al.~\cite{zhu2017event}, Gallego et al.~\cite{gallego2017accurate} and Rebecq et al.~\cite{rebecq2017real,rebecq2017emvs}, which took advantage of the high temporal resolution of the events to remove motion blur from an event image using an estimate of the motion in the scene, such as optical flow or camera pose. To estimate optical flow, we use the motion field equation, given camera velocity and a set of disparities, and similarly interpolate the position of the events at each disparity to a single point in time, which we represent as a novel temporally synchronized event disparity volume. We show that, in addition to removing motion blur at the correct disparities (where the motion field equation is valid), this volume allows us to disambiguate otherwise challenging regions in the image by inducing additional motion blur.

We then define a novel matching cost over this event disparity volume, which rewards similarity between patches, while penalizing blurriness inside the patch. We show that this cost function is able to robustly distinguish the true disparity, while being extremely cheap to compute, using only bitwise comparison operations over a sliding window.

Our method, implemented in Tensorflow, runs in realtime at 40ms on a laptop grade GPU, with significant further optimizations available. We evaluate our results on the Multi Vehicle Stereo Event Camera dataset\footnote{Dataset website: \url{https://daniilidis-group.github.io/mvsec/}}~\cite{zhu2018multi}, and show significant improvements in disparity error over state of the art event based stereo methods, which rely on additional, more computationally expensive, smoothness regularizations. 

The main contributions of this paper are summarized as:
\begin{itemize}
\item A novel method for using camera velocity to generate a time synchronized event disparity volume where regions at the correct disparity are in focus, while regions at the incorrect disparity are blurred.
\item A novel block matching based cost function over an event disparity volume that jointly rewards similarity between left and right patches while penalizing blurriness in both patches.
\item Evaluations on the Multi Vehicle Stereo Event Camera dataset, with comparisons against other state of the art methods, and evaluations of each component of the method.
\end{itemize}
\section{Related Work}
\label{sec:related_work}

Early works in stereo depth estimation for event cameras, such as the works by Kogler et al.~\cite{kogler2011event} and Rogister et al.~\cite{rogister2012asynchronous}, attempted to perform matching between individual events in a fully asynchronous fashion, using a combination of temporal and spatial constraints, which Kogler et al. showed to perform better than basic block matching between pairs of event images. However, these methods suffer from ambiguities in matching when single events are considered. 

To address these ambiguities, Camuñas-Mesa et al.~\cite{camunas2014use} use local spatial information in the form of local Gabor filters as features, while in \cite{camunas2017event}, they track clusters of events to aid in tracking with occlusion. Zou et al.~\cite{zou2016context} use a novel local event context descriptor based on the distances between events in a window, which they extend in \cite{zou2017robust} to produce a dense disparity estimate. Similarly, Schraml et al.~\cite{schraml2015event} use a cost based on the distance between events to generate panoramic depth maps.

In addition, several works have applied smoothing based regularizations to constrain ambiguous regions, which have seen great success in frame based stereo. Piatkowska et al.~\cite{piatkowska2014cooperative,piatkowska2017improved}, have applied cooperative stereo methods~\cite{marr1976cooperative} in an asynchronous fashion, while Xie et al.~\cite{xie2017event,xie2018event} have adapted belief propagation~\cite{besse2014pmbp} and semiglobal matching~\cite{felzenszwalb2006efficient}, respectively, to similar effect. These regularizations have shown significant improvements over the prior state of the art.

These prior works have all shown promising results for event-based stereo matching, but do not explicitly handle the time synchronization problem without abandoning the rich spatial information around each pixel. 

The problem of time synchronization has been approached in other problems, where it has been used to remove motion blur from event images. Zhu et al.~\cite{zhu2017event} and Gallego et al.~\cite{gallego2017accurate} use this synchronization as a cost function to estimate optical flow and angular velocity, respectively. Rebecq et al.~\cite{rebecq2017emvs} use the pose of a single camera from multiple views to generate a disparity space volume, in which the correct depth is similarly deblurred. Rebecq et al.~\cite{rebecq2017real} use a state estimator with pose and sparse depths to generate `motion compensated' event images, on which they perform feature tracking. More recently, Mitrokhin et al.~\cite{mitrokhin2018event} use the synchronized images to perform object detection and tracking.
\section{Method}
\label{sec:method}
The underlying problem of stereo disparity matching can be thought of as a data association problem. That is, to find correspondences between the points in the left and right images, at a given point in time. In this work, we assume that the cameras are calibrated and rectified, so that every epipolar line in both images is parallel to the x axis, and the correspondence problem is reduced to a 1D search along the x dimension. While some prior works such as \cite{rogister2012asynchronous} in the event based literature have tried to perform matching on an event by event basis, we use the spatial neighborhood around each pixel for a more detailed and robust matching, by making a locally constant depth assumption. 

It is possible to perform matching on event images generated directly from the event positions. However, such an image generated from the raw events is very prone to motion blur, unless the time window is carefully selected, with a method such as the lifetime estimation in \cite{mueggler2015lifetime}.

Motion blur is generated when events are captured at different points in time, such that events corresponding to the same point in the image may occur at different pixels due to the motion of that point. However, the works in \cite{zhu2017event}, \cite{gallego2017accurate} and \cite{rebecq2017real} show that motion blur can be removed from an event image if the optical flow for each pixel is known. In Sec. \ref{sec:time_shift}, we leverage this technique to both remove motion blur at the correct disparities, while further blurring the events at incorrect disparities. We then describe a novel event disparity volume representation of these time shifted events in Sec. \ref{sec:event_image}, on which we apply a novel cost function that leverages this focus-defocus effect to allow us to discriminate the true disparity at each pixel, as described in Sec. \ref{sec:matching_cost}. Finally, Sec. \ref{sec:disparity_est} discusses methods to then use the cost function to estimate the true disparity at each pixel.

\begin{figure}[t!]
\centering
\includegraphics[width=0.98\linewidth]{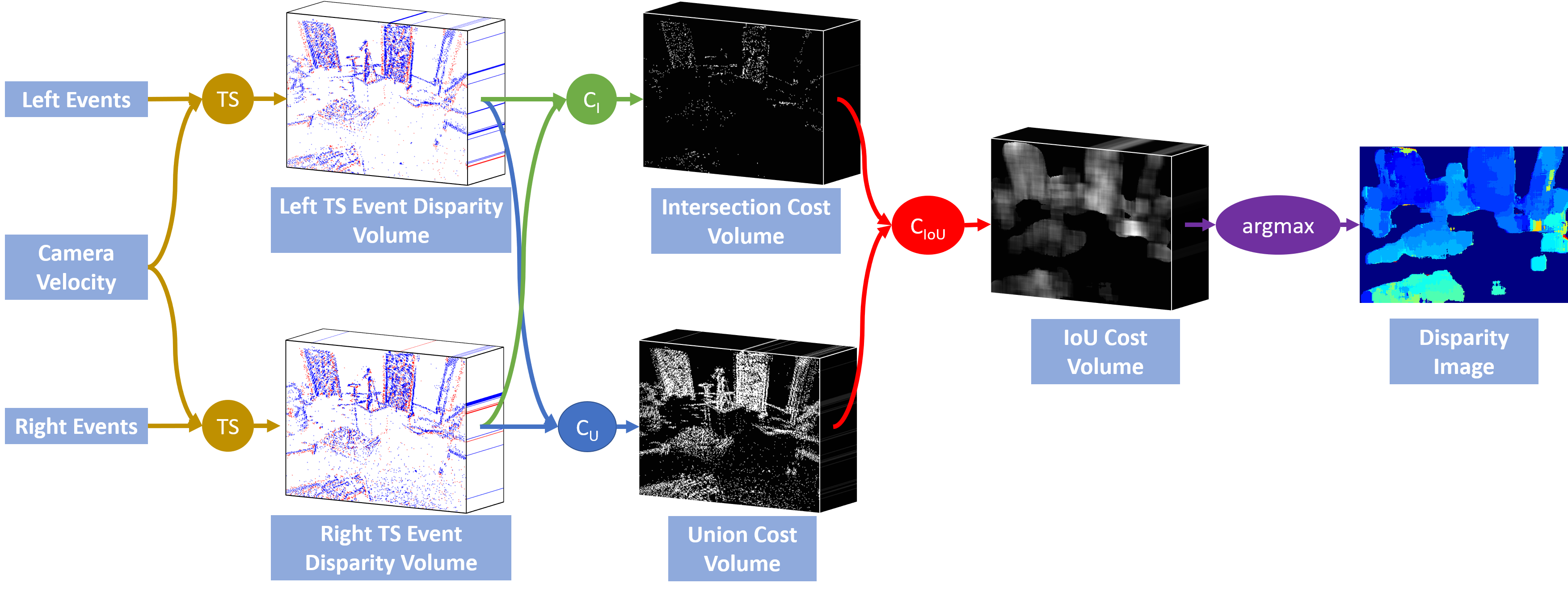}  
  \caption{Overview of our method. Given an input of left and right events and camera velocity, left and right time synchronized event disparity volumes are generated (Sec. \ref{sec:time_shift} and \ref{sec:event_image}). The intersection and union costs are calculated by combining the two disparity volumes, and the final IoU cost volume is computed (Sec. \ref{sec:matching_cost}). Finally, the disparity is computed in a winner takes all scheme over the IoU cost volume (Sec. \ref{sec:disparity_est}). Best viewed in color.}
  \label{fig:method}
\end{figure}

An overview of the method can be found in Fig. \ref{fig:method}

\subsection{Time Synchronization through Interpolation}
\label{sec:time_shift}
For a given disparity, $d$, we can approximate optical flow using the motion field equation, with an assumption of known camera velocity. The motion field equation describes the relationship between the linear ($\mathbf{v}$) and angular ($\boldsymbol{\omega}$) velocity of a camera, depth $Z$ of a point $(x, y)$, which we treat here as a function of disparity, $d$, and the motion of the point in the image, which we approximate to be the optical flow ($\dot{x}_i$, $\dot{y}_i$):
\begin{align}
\begin{pmatrix}\dot{x}_i(d)\\\dot{y}_i(d)\end{pmatrix}=&\frac{1}{Z(d)}\begin{bmatrix}-1 & 0 & x_i\\ 0 & -1 & y_i\end{bmatrix}\mathbf{v}+\begin{bmatrix} x_iy_i & -(1+x_i^2) & y_i\\ 1+y_i^2 & -x_iy_i & -x_i\end{bmatrix}\boldsymbol{\omega}\\
Z(d)=&\frac{fb}{d}\label{eq:motion_field}
\end{align}
where $f$ is the focal length of the camera and b is the baseline between the two cameras.

Assuming that the optical flow for each pixel is constant within each time window, we can then estimate the position of a point generating an event $(x_i,y_i,t_i,p_i)$ at a constant time $t'$ with linear interpolation:
\begin{align}
\begin{pmatrix}x_i'(d)\\y_i'(d)\end{pmatrix}=&\begin{pmatrix}x_i\\y_i\end{pmatrix}+\begin{pmatrix}\dot{x}_i(d)\\\dot{y}_i(d)\end{pmatrix}(t'-t_i)\label{eq:time_shift}
\end{align}
Assuming a static scene and accurate velocities and disparities, the set of time synchronized events, $\left\{\begin{pmatrix}x_i'(d)& y_i'(d) & t' & p_i\end{pmatrix}\right\}$, is assumed to have no motion blur for all $x_i$ and $y_i$ with disparity $d$.

\begin{figure}[t!]
\centering
\begin{subfigure}[b]{0.98\textwidth}
\includegraphics[width=\linewidth]{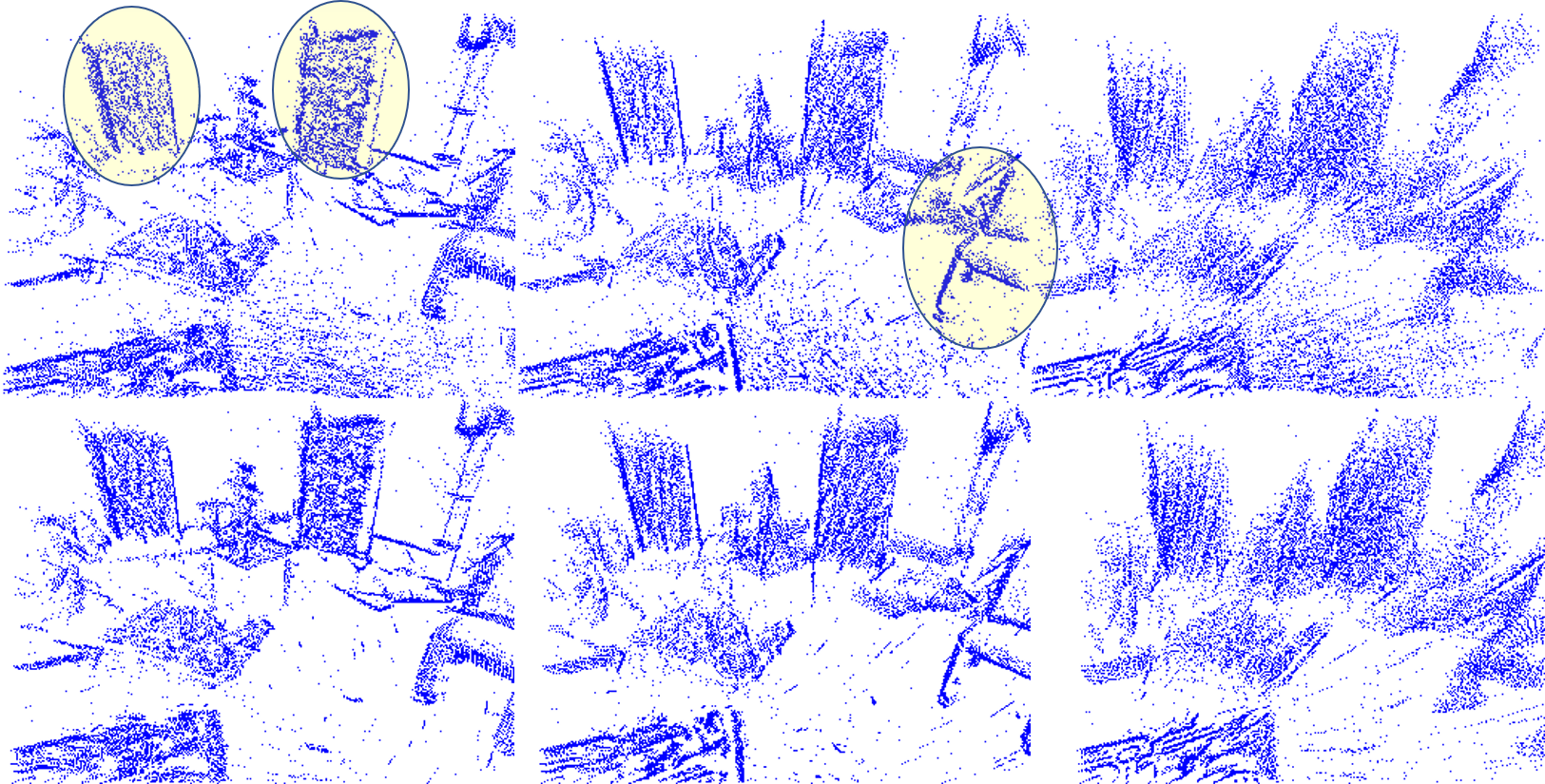}
\caption*{Disparity = 6\hspace{0.17\textwidth}Disparity = 16\hspace{0.17\textwidth}Disparity = 31}
\end{subfigure}
  \caption{Sample slices of the left (top) and right (bottom) time synchronized event disparity volumes at disparities 6 (left), 16 (middle) and 31 (right). Only positive pixels are shown for clarity. At disparity 6, the boards at the back are in focus, while at disparity 16, the chair in the front is in focus (both circled in yellow). The other features at the wrong depth are blurred. The right slices have been shifted horizontally by the disparity, as in \eqref{eq:right_event_image}, so that corresponding points should be at the same x position in both images. Best viewed in color.}
  \label{fig:disparity_volume}
\end{figure}

\subsection{Time Synchronized Event Disparity Volume Generation}
\label{sec:event_image}
However, the true depth for each pixel is unknown for this problem. Instead, we select a range of disparities over which to search, and apply \eqref{eq:time_shift} to the set of events from the left camera for every disparity within the range.

At each disparity level, $d$, we generate an image based on the time shifted events, where a pixel with more positive events is set to 1, more negative events is set to -1, and no events is set to 0. Note that the time shifted event positions are rounded to the nearest integer to index into the image.
\begin{align}
I_{L}(x,y,d)=&\;\text{sign}\left(\sum_{i}p_i\right)\label{eq:left_event_image}\\
i\in& \{i|(x'_{i}(d), y'_{i}(d))=(x, y)\}\nonumber\\
p_i\in&\{-1, 1\}\nonumber
\end{align}
This is similar to standard methods that generate images by summing events at each pixel, but the additional sign operator allows the image to be robust to the left or right camera generating more events at each pixel than the other.

The result is a 3D volume for the left camera, where each slice in the disparity dimension represents the images generated according to \eqref{eq:left_event_image}, using the disparity corresponding to that slice. That is, when the camera moves with some linear velocity, this flow would have a deblurring effect on points where the pixel position matches the disparity, and potentially apply further blurring on points where the disparity is incorrect. In the case when the camera's motion is pure rotation, the flow will produce unblurred images at each disparity slice.

We apply a similar operation to the events from the right camera, except that the x position of each shifted event is further shifted by the disparity at each level:
\begin{align}
I_{R}(x,y,d)=&\;\text{sign}\left(\sum_{i}p_i\right)\label{eq:right_event_image}\\
i\in& \{i | (x'_{i}(d)+d, y'_{i}(d))=(x, y)\}\nonumber\\
p_i\in&\{-1, 1\}\nonumber
\end{align}
This generates a set of disparity volumes similar to traditional plane sweep volumes, where the potential matching right pixel corresponding to $I_{L}(x, y, d)$ is $I_{R}(x, y, d)$. We show some example slices of this volume in Fig. \ref{fig:disparities}, where the blurring and deblurring effects can be clearly seen.

\subsection{Matching Cost}
\label{sec:matching_cost}

Finally, we apply a novel sliding window matching cost that leverages both the deblurring and blurring effects of Sec. \ref{sec:time_shift}. First, it penalizes windows with many events, as this would indicate areas with an incorrect disparity due to the blurring incurred by the temporal interpolation. Given a local spatial window $W(x, y, d)$ around a pixel $(x, y)$ at a given disparity $d$, we encode this using a union term, defined as:
\begin{align}
C_{U}(x, y, d)=&\sum_{x^*, y^*\in W(x,y,d)} I_{L}(x^*, y^*, d) \cup I_{R}(x^*, y^*, d) \label{eq:union_cost}\\
a\cup b=&\left\{\begin{array}{ccc}1 & & a\neq 0 \text{ or } b \neq 0 \\ 0 & & \text{otherwise}\end{array}\right.\nonumber
\end{align}
We carefully choose the union operator instead of addition, in order to not double penalize pixels with events in both volumes.

Second, the cost rewards windows that are similar. That is, we would like pixels between the two images to have events with the same polarity. We encode this using an intersection term, defined as:
\begin{align}
C_{I}(x, y, d) =& \sum_{x^*, y^*\in W(x,y,d)} I_{L}(x^*, y^*, d) \cap I_{R}(x^*, y^*, d) \label{eq:intersection_cost}\\
a\cap b=&\left\{\begin{array}{ccc}1 & & a=b\neq 0 \\ 0 & & \text{otherwise}\end{array}\right.\nonumber
\end{align}
This is similar to the tri-state logic error function presented in \cite{kogler2011event}, except we explicitly do not reward pixels that are both 0 in the intersection term, as we only want to capture associations between events, and not between pixels without events.

The final cost, then, can be thought of as an analogy to the intersection over union cost:
\begin{align}
C_{IoU}(x, y, d)=&-\frac{C_{I}(x, y, d)}{C_{U}(x, y, d)}\label{eq:total_cost}
\end{align}
Minimizing this final cost will implicitly maximize the similarity between the two windows, while minimizing the blurring in each. By computing this cost function at every pixel and disparity, we generate a cost volume, where each element $(x,y,d)$ in the volume contains the cost of pixel $(x,y)$ being at disparity $d$. 

\subsection{Disparity Estimation}
\label{sec:disparity_est}
Given the cost volume, the fastest way to obtain the estimate for the true disparity at each pixel is to compute the argmax across the disparity dimension of the cost volume, in a winner takes all fashion:
\begin{align}
\hat{d}(x, y) = \argmin_{d} \; C_{IoU}(x, y, d)
\end{align}
However, we can also apply any traditional optimization method for stereo disparity estimation over the cost volume, such as semi-global matching \cite{felzenszwalb2006efficient} or belief propagation \cite{besse2014pmbp}.

\subsection{Outlier Rejection}
\label{sec:outliers}
While the cost function in \eqref{eq:total_cost} was relatively robust in our experiments, there were still some regions of the image where it was unable to resolve the correct disparity, which we need to remove from the final output. In particular, we found that pixels with a low final IoU cost typically corresponded to pure noise in the image, where the number of intersection matches was low compared to the number of events in the window. Therefore, any disparities with $C_{IoU}$ less than a parameter $\epsilon_c$ are considered outliers. In addition, windows with a low number of events do not provide enough support to find a meaningful match, and so we consider outliers any disparities with $C_{U}$ less than $\epsilon_n\times \|W\|$, where $\epsilon_n$ is a parameter and $\|W\|$ is the number of pixels in the spatial neighborhood.
\section{Implementation Details}
In our experiments, unless otherwise stated, we use a disparity range ranging from 0 to 31 pixels, and a square window with side length of 24 pixels. For outlier rejection, $\epsilon_c$ and $\epsilon_n$ were both set to 0.1. At each time step, a constant number of events is passed to the algorithm. For our experiments, we used 15,000 events. 

As every step of the algorithm is vectorizable with matrix notation, the algorithm was efficiently implemented on GPU in Tensorflow. In particular, \eqref{eq:motion_field} and \eqref{eq:time_shift} are implemented as a matrix operations, \eqref{eq:left_event_image} and \eqref{eq:right_event_image} are performed using scatter$\_$nd, and the costs in \eqref{eq:union_cost} and \eqref{eq:intersection_cost} are computed by computing the costs for each pixel at each disparity, and applying two 1D depthwise convolutions with a kernel of ones of the same length as the window size (one along the rows, one along the columns). 

With all operations fully vectorized, the algorithm takes 40ms to run on a laptop NVIDIA 960M GPU, including transfer time to the GPU. With further optimizations and an implementation in raw CUDA or OpenCL, we expect this time to reduce further. This corresponds to a runtime of around 2.7$\mu$s per event, compared to the 0.65-2ms reported in \cite{xie2017event}. However, it should be noted that the competing methods were implemented in MATLAB on CPU, and would almost certainly see speed improvements if ported to other languages/devices. In addition, our method is relatively insensitive to the number of events, as a large proportion of the run time ($\sim$40\%) is consumed in the sliding window cost. For example, processing a window of 30,000 events takes 46ms to run, corresponding to a runtime of 1.53$\mu$s per event.
\begin{figure}[h!]
\centering
\includegraphics[width=0.19\linewidth]{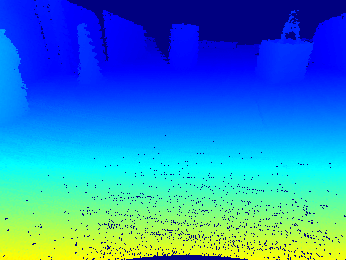}
\includegraphics[width=0.19\linewidth]{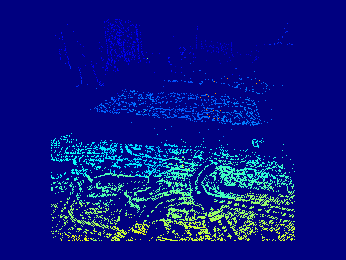}  
\includegraphics[width=0.19\linewidth]{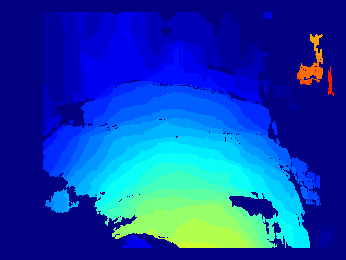}
\includegraphics[width=0.19\linewidth]{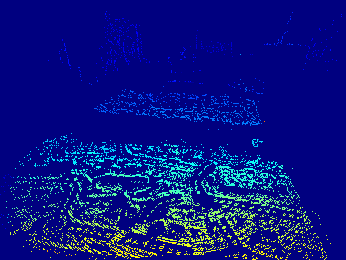}  
\includegraphics[width=0.19\linewidth]{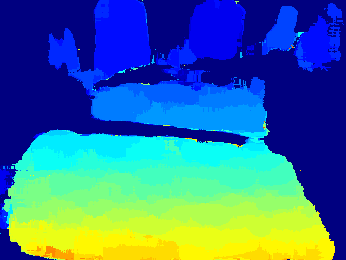}

\includegraphics[width=0.19\linewidth]{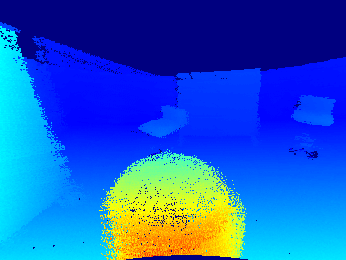}
\includegraphics[width=0.19\linewidth]{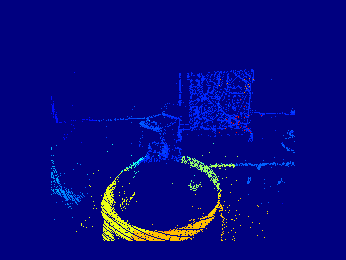}
\includegraphics[width=0.19\linewidth]{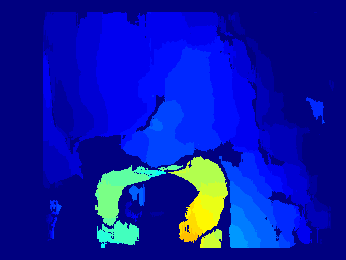}
\includegraphics[width=0.19\linewidth]{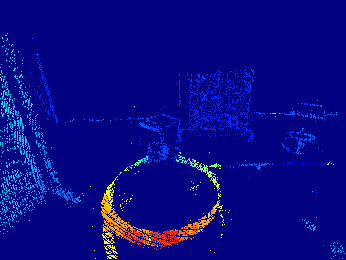}  
\includegraphics[width=0.19\linewidth]{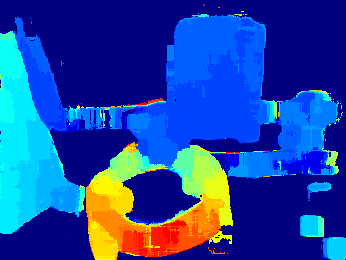}

\includegraphics[width=0.19\linewidth]{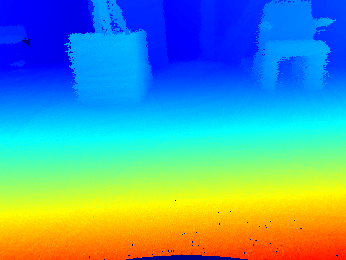}
\includegraphics[width=0.19\linewidth]{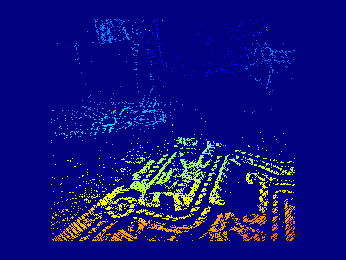}  
\includegraphics[width=0.19\linewidth]{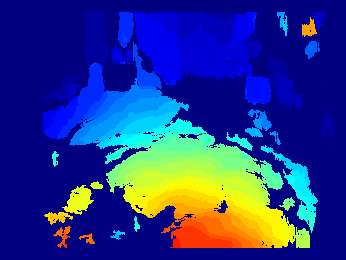}
\includegraphics[width=0.19\linewidth]{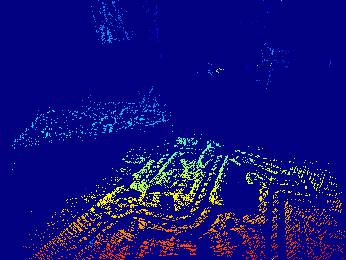}  
\includegraphics[width=0.19\linewidth]{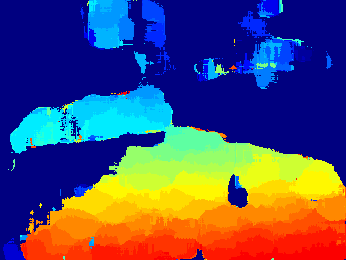}

\includegraphics[width=0.19\linewidth]{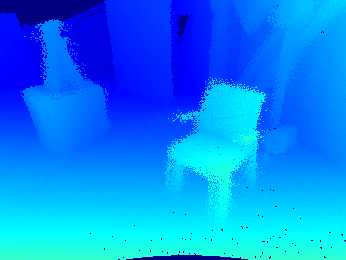}
\includegraphics[width=0.19\linewidth]{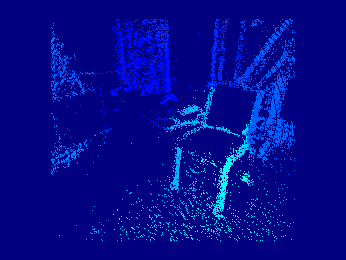}  
\includegraphics[width=0.19\linewidth]{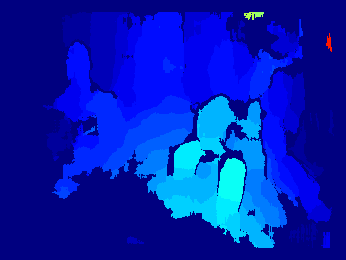}
\includegraphics[width=0.19\linewidth]{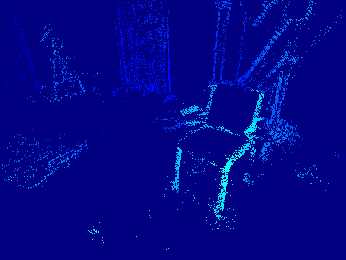}  
\includegraphics[width=0.19\linewidth]{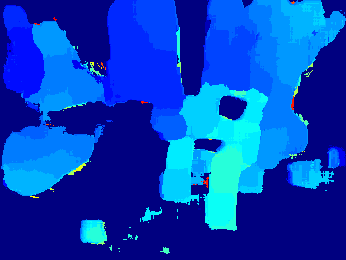}  

\begin{subfigure}[b]{0.19\textwidth}
\includegraphics[width=\linewidth]{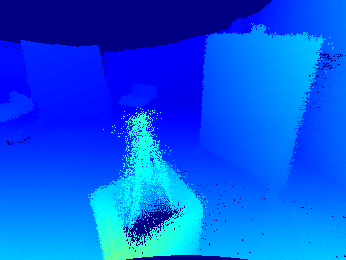}
\caption*{Ground Truth}
\end{subfigure}
\begin{subfigure}[b]{0.19\textwidth}
\includegraphics[width=\linewidth]{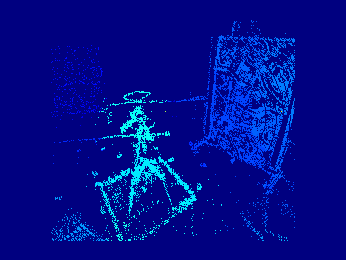}
\caption*{CopNet}
\end{subfigure}
\begin{subfigure}[b]{0.19\textwidth}
\includegraphics[width=\linewidth]{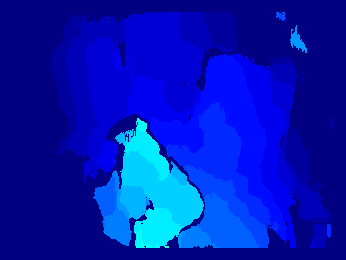}
\caption*{Block Matching}
\end{subfigure}
\begin{subfigure}[b]{0.19\textwidth}
\includegraphics[width=\linewidth]{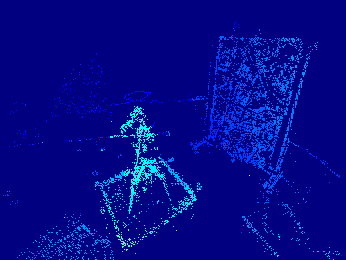}
\caption*{TSES Sparse}
\end{subfigure}
\begin{subfigure}[b]{0.19\textwidth}
\includegraphics[width=\linewidth]{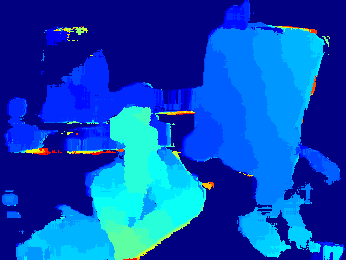}
\caption*{TSES Dense}
\end{subfigure}
  \caption{Sample outputs from TSES (our method), compared against CopNet and block matching, with ground truth from MVSEC. Pixels without disparities are dark blue. Note that the border of the CopNet and block matching results are empty due to the window size. Quantitative results were only computed over points with disparities. Best viewed in color.}
  \label{fig:disparities}
\end{figure}
\begin{table}[h!]
\centering
\begin{tabular}{C{2.5cm}C{1cm}C{1cm}C{1cm}C{1cm}C{1cm}C{1cm}C{1cm}C{1cm}c}
\toprule
 & \multicolumn{3}{c}{Mean Disp. Error (pix)} & \multicolumn{3}{c}{Mean Depth Error (m)} & \multicolumn{3}{c}{\% Disp. Err $<$ 1} \\ 
\cmidrule(lr){2-4}\cmidrule(lr){5-7}\cmidrule(lr){8-10}
 & IF1 & IF2 & IF3 & IF1 & IF2 & IF3 & IF1 & IF2 & IF3 \\
\hline
TSES & 
0.89			& 1.98 			& 0.88			& 
0.36 			& 0.44			& 0.36			& 
\textbf{82.3} 	& \textbf{70.1} & \textbf{82.3} \\
CopNet & 1.03 & 1.54 & 1.01 & 0.61 & 1.00 & 0.64 & 70.4 & 52.8 & 70.6\\
BM & 
\textbf{0.73} 	& \textbf{1.02} & \textbf{0.82} & 
\textbf{0.23} 	& \textbf{0.21} & \textbf{0.27} & 
79.5		 	& 65.2			& 74.3\\
SGBM & 
1.96 & 3.06 & 1.86 &
0.38 & 0.38 & 0.41 &
69.9 & 56.8 & 66.7\\
\hline
\multicolumn{10}{c}{Algorithm Ablation}\\
\hline
T-S & 1.30 & 2.54 & 1.39 & 0.50 & 0.58 & 0.57 & 77.3 & 64.9 & 76.7 \\
I-S & 1.71 & 3.59 & 1.99 & 0.67 & 0.99 & 0.77 & 74.2 & 60.7 & 72.5 \\
IoU-NS & 1.43 & 2.29 & 1.42 & 0.52 & 0.47 & 0.53 & 67.8 & 59.0 & 68.3 \\
T-NS & 1.85 & 2.78 & 1.84 & 0.76 & 0.66 & 0.78 & 64.2 & 55.6 & 64.0 \\
I-NS & 2.21 & 3.20 & 2.12 & 0.80 & 0.80 & 0.78 & 61.6 & 53.5 & 62.7 \\
TSES w/ outliers &
1.87 			& 2.83 			& 1.73 			&
1.28			& 1.18			& 1.15			&
74.3			& 64.3			& 75.2\\
\hline
\multicolumn{10}{c}{Velocity Noise Ablation}\\
\hline
0\% & 
\textbf{0.89}	& 1.98 			& \textbf{0.88}	& 
\textbf{0.36}	& \textbf{0.44}	& \textbf{0.36}	& 
\textbf{82.3}	& 70.1			& 82.3          \\
5\% &
0.90 			& \textbf{1.97}	& \textbf{0.88}	&
\textbf{0.36}	& 0.45	& \textbf{0.36}	&
82.0			& \textbf{70.5}	& \textbf{82.4} \\
10\% &
0.91			& 1.98			& \textbf{0.88}	&
0.37	        & 0.45	& \textbf{0.36}	&
81.6			& 70.1			& 82.3          \\
20\% & 
0.96 			& 2.04			& 0.92			&
0.38			& 0.46			& 0.38			&
80.4			& 68.6			& 81.3			\\
50\% & 
1.21 			& 2.44 			& 1.23 			& 
0.47 			& 0.58 			& 0.51 			& 
74.5 			& 61.5			& 74.5			\\
100\% & 
1.97 			& 3.47 			& 2.17 			& 
0.83 			& 0.92 			& 1.03 			&
61.8 			& 48.5 			& 59.1\\
\hline
\multicolumn{10}{c}{Window Size Ablation}\\
\hline
8 pix. & 
2.40 			& 3.83 			& 2.52 			& 
0.78 			& 0.87 			& 0.86 			& 
65.4 			& 55.3 			& 63.8\\
16 pix. & 
1.10			& 2.29 			& 1.13 			& 
0.43 			& 0.51 			& 0.45 			& 
80.3 			& 69.2 			& 79.8\\
24 pix. & 
0.89 			& 1.98          & 0.88 			& 
0.36 			& 0.44          & 0.36 			& 
\textbf{82.3}   & \textbf{70.1}	& \textbf{82.3}\\
32 pix. & 
\textbf{0.86} 	& \textbf{1.97} & \textbf{0.84}	& 
\textbf{0.34}	& \textbf{0.43} & 0.34 			& 
81.2 			& 66.7 			& 81.4\\
40 pix. & 
0.89			& 2.05 			& 0.91 			& 
\textbf{0.34} & 0.44 			& \textbf{0.33} & 
78.6 			& 61.9 			& 77.9\\
\hline\\
\end{tabular}
\caption{Quantitative results from testing on the indoor$\_$flying (IF) sequences of TSES (our method) and CopNet, along with ablation studies. Prefixes for the algorithm ablation are: IoU - Intersection over Union cost \eqref{eq:total_cost}, I - Intersection cost \eqref{eq:intersection_cost}, T - Time cost \eqref{eq:time_cost}. Suffixes are with (S) and without (NS) time synchronization \eqref{eq:time_shift}. Velocity noise was added to the linear and angular velocities separately, as zero mean Gaussian noise with variance equal to a percentage of the norm of each velocity.}
\label{tab:comparison}
\end{table}
\section{Experiments}
\subsection{Data}
We evaluated our algorithm on the Multi Vehicle Stereo Event Camera (MVSEC) dataset~\cite{zhu2018multi}. MVSEC provides data captured from a stereo event camera pair, along with grayscale images and ground truth depth and pose of the cameras. We tested our method on the indoor$\_$flying sequences, and evaluated against the provided ground truth depth maps. These sequences were generated from a stereo event camera pair mounted on a hexacopter, and flown in an indoor environment, with ground truth generated from lidar measurements. In particular, we used the following depth map frames (zero index) from these indoor$\_$ flying sequences for evaluation:
indoor$\_$flying1: 140-1200,
indoor$\_$flying2: 120-1420,
indoor$\_$flying3: 73-1616. These frames were selected to exclude the takeoff and landing frames where the ground is closer than our selected maximum disparity. 

The driving sequences were not included as the majority of the points in those sequences were beyond the depth resolved by a disparity of 1, and so a sub-pixel disparity estimator would be needed to achieve accurate results. In addition, we do not include results from indoor$\_$flying4, as the majority of events are closer than the maximum disparity of 31, and are also generated by the low-texture floor, on which we could not generate reasonable results with any of the methods. 

While our method generates disparity values whenever there are any events inside the spatial window, we report our results based on disparities on pixels where events appeared, in order to provide a fair comparison with other works.

We used the camera velocities provided in the dataset from \cite{zhu2018ev}, which were generated by linear interpolation of the lidar odometry poses provided from MVSEC, and are provided in addition to ground truth optical flow for the sequences in the dataset.

\subsection{Comparisons}
For comparison, we have implemented the CopNet method by~Piatkowska et al. \cite{piatkowska2017improved}, and we include their results on the same dataset, using their provided parameters. For these experiments, we have used an $\alpha$ value of 1 (the scaling term in the matching cost, equation (3) in their paper), as the original paper stated a value of 0, which would result in a constant cost. In addition, we compare against block matching and semi-global block matching methods from OpenCV\footnote{\url{https://docs.opencv.org/3.4/d2/d6e/classcv_1_1StereoMatcher.html}}, applied to the grayscale frames from the DAVIS camera. Note that the grayscale frames are not time synchronized, and the time offset between the left and right frames is 4ms, 14ms and 14ms for indoor$\_$flying 1, 2 and 3, respectively. However, we were still able to achieve reasonable performance. The quantitative results of these comparisons can be found in Tab. \ref{tab:comparison}.

In addition, we attempted an implementation of the belief propagation based work by Xie et al.~\cite{xie2017event}, but were unable to obtain reasonable results over this dataset, which is significantly more complex than those evaluated in the original work, consisting of a few objects moving in the scene. We believe that this is because their matching cost ($D(d_p)$) attempts to match individual events, without using the spatial neighborhood around the event. In our experiments, this matching cost failed to identify the correct disparity over the majority of the image, which we believe led the belief propagation to output incorrect results.

\subsection{Ablation Studies}
In addition to the comparisons, we performed a number of ablation studies over the parameters of the algorithm. All results can be found in Tab. \ref{tab:comparison}.
\subsubsection{Algorithm Ablation}
To test the effect of the time synchronized event disparity volumes, we performed additional experiments where the raw event positions were passed directly into \eqref{eq:left_event_image} and \eqref{eq:right_event_image} (i.e. by setting $(x'(d)_i, y'(d)_i)=(x_i, y_i)$). Experiments with and without time synchronization are denoted with the suffix -S and -NS, respectively.

To test the IoU cost, we tested with only the intersection cost (prefix I), as well as using the cost function from \cite{piatkowska2017improved} (prefix T), which is defined as:
\begin{align}
C_{T}(x,y,d)=&\sum_{x^*, y^*\in W(x,y,d)}\frac{1}{(\alpha\cdot |I^t_{L}(x^*,y^*,d)-I^{t}_{R}(x^*,y^*,d)| + 1)\cdot C_{U}(x^*, y^*, d)} \label{eq:time_cost}
\end{align}
where $\alpha$ is set to 1, the event images $I^{t}$ now represent the timestamp of the last event to arrive at each pixel and disparity:
\begin{align}
I^{t}_{L}(x, y, d)=&\max_{t_i} \; \{t_i|(x'_i(d), y'_i(d)) = (x, y)\}\\
I^{t}_{R}(x, y, d)=&\max_{t_i} \; \{t_i|(x'_i(d)+d, y'_i(d)) = (x, y)\}
\end{align}
and we use our union cost in place of the number of events in the left window. We also tested with the inverse of the union cost, but this did not produce any reasonable results.

We also provide results of our full method, without the outlier rejection step.

\subsubsection{Velocity Noise Ablation}
In practice, it is difficult to estimate the cameras' velocity with the same accuracy as the ground truth. To test the effect of noise on the velocity estimate, we perform additional experiments where we add zero mean Gaussian noise to the linear and angular velocities. The variance of the noise is set to a given percentage of the norm of the linear and angular velocities, separately.

\subsubsection{Window Size Ablation}
We also tested the effects of the window size on the performance of our method over a range of window sizes.

\subsection{Results and Discussion}
\subsubsection{Comparisons}
We present some qualitative results in Fig. \ref{fig:disparities}, comparing our method to CopNet, block matching and ground truth. While both sets of results look visually reasonable, we can see that our method suffers less from foreground fattening \cite{geiger2010efficient} (e.g. the chair in the fourth row). Our method does, however, tend to produce erroneous results on the edges of images in the dense disparity image, but these correspond to pixels without any events, and are thresholded away in the sparse disparity image.

In addition, we provide quantitative results in Tab. \ref{tab:comparison}, where we can see that our full method outperforms CopNet across almost every measure, as well as the other methods in the ablation study. In particular, while the disparity errors are similar, CopNet performs significantly worse in depth overall. Upon examining the results. We found that this error from CopNet was largely due to the fact that the method had over-smoothed the disparity output. This was mostly due to the window size used, which is relatively large (39x39). This oversmoothing tends to pull far away points closer (overestimates disparities), which leads to higher depth errors, as they are higher at lower disparity levels. However, we observed that reducing the window sizes resulted in a further reduction in the overall matching accuracy due to increased ambiguity in the matching, as noted by the authors in the original paper \cite{piatkowska2017improved}, so there was no immediate solution for this problem.

The block matching method performed better in terms of mean errors across all three sequences, although the mean disparity errors for sequences 1 and 3 are both less than 1 pixel, which is within the range of the discretization error. In addition, our method has a higher percentage of points with disparity error $<$1 across all sequences.

We were unable to achieve comparable performance from semi-global block matching, which tended to over smooth incorrect regions in the image.

\subsubsection{Algorithm Ablations}
From the ablation study, we can see that removing each component of the method tends to result in a corresponding decrease in accuracy, with the time synchronization always resulting in better results. In addition, the addition of the union cost to the overall cost provides a significant improvement in accuracy over the intersection cost, which is a pure similarity measure. 

Furthermore, we can see that our IoU cost outperforms the timestamp based cost in both situations, suggesting that it may be a better alternative for more complex methods, even without the time synchronization. When the proposed time synchronization is applied at the correct disparity, older timestamps are mapped to later timestamps from the same point in the image. As the time cost operates on an image that only keeps the latest timestamp, this results in images with timestamps that are very similar (all later events), which do not provide much discriminative power. Future work could consider all of the events that map to a pixel, but this requires a new cost function.

Finally, the results without outlier rejection have significantly higher mean disparity errors, suggesting that a large number of outliers were rejected by our method, while from the \% disparity error $<$ 1 results, we can see that only $<$10\% of the points were rejected.

\subsubsection{Velocity Noise Ablation}
The velocity noise ablation results show stable errors up to noise with variance up to around 20\% of the velocity norm. We believe that a conventional state estimation pipeline for event cameras should be able to reliably estimate the camera velocity within these error bounds.

\subsubsection{Window Size Ablation}
We found that window sizes between 24 and 40 pixels achieved the best results. However, larger window sizes increase the amount of foreground fattening, as well as the runtime of the algorithm. Therefore, we recommend a window size of 24 pixels for these test cases. In terms of run time, the algorithm took 33ms, 40ms and 50ms to run for window sizes of 16, 24 and 32 pixels, respectively. Similarly, the runtime was 25ms and 60ms for disparity ranges of 16 and 48, with a window size of 24 pixels.

\section{Conclusions}
\label{sec:conclusions}
We have proposed a novel method for stereo event disparity matching which uses the motion of the camera to synchronize the event streams in time. We show that our method, consisting of a simple temporal interpolation of the events, along with a lightweight matching cost, is able to outperform state of the art methods which perform expensive regularizations on top of the disparity map. In addition, as our disparity results are at a single time, analogous to an image frame, they can be directly passed into any frame based architecture such as a state estimator, as compared to an asynchronous disparity stream. We envision that this method will be coupled with a method for estimating camera velocity, such as a visual odometry algorithm, for real time performance.
\section*{Acknowledgements}
Thanks to Tobi Delbruck and the team at iniLabs for providing and supporting the DAVIS-346b cameras. We also gratefully appreciate support through the following grants: NSF-IIS-1703319, NSF-IIP-1439681 (I/UCRC),  ARL RCTA W911NF-10-2-0016, and the DARPA FLA program.
\bibliographystyle{splncs04}
\bibliography{related_work}
\end{document}